\documentclass[lettersize,journal]{IEEEtran}
\usepackage{amsmath,amsfonts}
\usepackage{algorithm}
\usepackage{array}
\usepackage[caption=false,font=normalsize,labelfont=sf,textfont=sf]{subfig}
\usepackage{textcomp}
\usepackage{stfloats}
\usepackage{url}
\usepackage{orcidlink}
\usepackage{verbatim}
\usepackage{graphicx}
\usepackage{cite}
\usepackage{tabularx}
\usepackage{booktabs}
\usepackage{amsthm}
\newtheorem{example}{Example}
\usepackage{algpseudocode}
\theoremstyle{definition}
\newtheorem{definition}{Definition}
\usepackage{xcolor}
\newcommand{\inconsistent}[1]{\colorbox{red!20}{#1}}
\newcommand{\consistent}[1]{\colorbox{green!20}{#1}}
\newcommand{\completed}[1]{\colorbox{blue!15}{#1}}
\usepackage{url}
\begin{document}

\title{ProCap: Prominence-guided Object Rectification\\for Faithful and Comprehensive Video Captioning} 


\author{
Debjyoti Das Adhikary\orcidlink{0009-0004-3086-4501},
Aritra Hazra\orcidlink{0000-0003-2076-3577},
and Partha Pratim Chakrabarti\orcidlink{0000-0002-3553-8834},%

\thanks{
The authors are with the Department of Computer Science and Engineering,
Indian Institute of Technology Kharagpur, Kharagpur 721302, India
(e-mail:
debjyoti.das.adhikary@kgpian.iitkgp.ac.in;
aritrah@cse.iitkgp.ac.in;
ppchak@cse.iitkgp.ac.in).}
}


\markboth{Journal of \LaTeX\ Class Files,~Vol.~14, No.~8, August~2021}%
{Shell \MakeLowercase{\textit{et al.}}: A Sample Article Using IEEEtran.cls for IEEE Journals}


\maketitle

\begin{abstract}
Improving video captioning quality typically demands retraining large vision-language models, an expensive and often impractical requirement. Existing training-free alternatives instead ground captions in detected objects to curb hallucination, but apply only a single, fixed correction pass without prioritizing which objects matter most, leaving semantically significant content omitted. We propose a prominence-aware, iterative post-hoc rectification framework that overcomes both limitations without modifying the underlying captioning model's parameters: a lightweight scoring mechanism ranks detected objects by spatial saliency, temporal persistence, and relational dynamics, and an iterative, prompt-driven refinement loop uses this ranking to progressively inject missing yet contextually relevant objects into the caption over multiple rounds. We validate the framework on MSVD and MSR-VTT using object-grounded automatic metrics, a 110-participant human study, and qualitative comparison against ChatGPT and Gemini; in human evaluation, the framework raises perceived completeness by up to 48\% and reduces hallucination by up to 45\% relative to a strong pretrained captioning baseline, all without retraining or reference captions. These results position prominence-guided iterative rectification as a lightweight, scalable, and model-agnostic route to more complete and trustworthy video captioning, with direct relevance to accessibility, retrieval, and other multimedia understanding applications.
\end{abstract}


\section{Introduction}
Recent advances in Vision Language Models (VLMs) and Large Language Models (LLMs) have driven substantial progress in multimodal understanding, spanning image captioning, visual question answering, visual grounding, and video understanding. Among these, video captioning -- the task of generating a concise, semantically faithful natural-language description of a video -- has emerged as a key enabling technology for accessibility, video retrieval and summarization, surveillance, and Human Computer Interaction~\cite{liu2021makes, wu2025video, tellex2020robots}.

Unlike image captioning, video captioning requires jointly modeling spatial appearance, temporal evolution, and object interactions across a sequence of frames while remaining faithful to the visual content, without hallucinating or omitting salient entities -- a far from academic concern, since overlooking a critical object in a surveillance clip, hallucinating entities in a medical video, or producing an ambiguous accessibility caption can each directly mislead downstream decisions. Effective video captioning therefore demands joint reasoning over appearance, motion, and temporal consistency, delivering semantic completeness and visual faithfulness alongside linguistic fluency.

Driven by large-scale pretraining, the field has shifted from CNN/RNN encoder--decoder pipelines toward Transformer-based vision--language foundation models such as mPLUG-2~\cite{xu2023mplug}, Video-LLaMA~\cite{zhang2023video}, and Gemini~\cite{comanici2025gemini}, which substantially improve fluency and zero-shot generalization~\cite{yuan2023gpt} but remain fundamentally generative -- prioritizing linguistic plausibility over strict visual grounding, and consequently prone to omitting or hallucinating objects in ways standard reference-based metrics (BLEU, METEOR, CIDEr) do not penalize. Post-hoc rectification methods address this without retraining by grounding captions in externally detected objects; our own prior framework, ReCap, illustrates their key limitation, however -- it corrects captions only at the frame level and aggregates them via a single-pass LLM summarization, treating every detected object as equally important and leaving semantically significant content omitted. Section~\ref{sec:related_work} discusses this literature in detail.

This motivates a framework that (i) explicitly ranks detected objects by semantic importance rather than treating them uniformly, and (ii) refines the video-level caption iteratively rather than in a single pass -- entirely post-hoc, without modifying the underlying captioning model itself.

\begin{figure}[!t]
    \centering
    \includegraphics[width=\columnwidth]{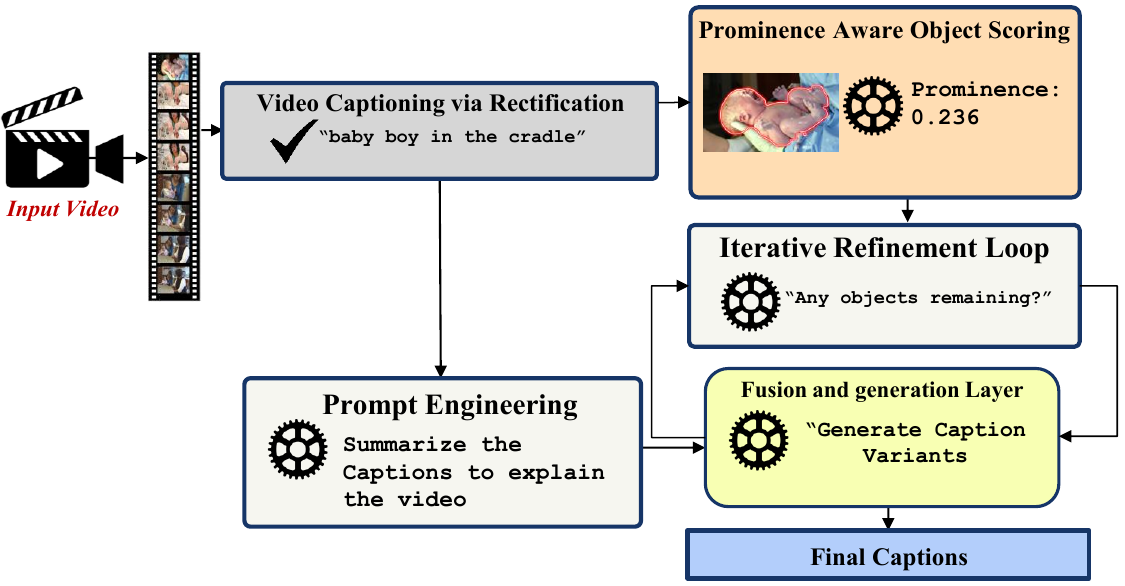}
    \caption{Overview of the proposed prominence-guided rectification framework.}
    \label{fig:teaser_figure}
\end{figure}

We introduce a prominence-aware, iterative post-hoc rectification framework that extends ReCap along both axes (Fig.~\ref{fig:teaser_figure}). A prominence score jointly models each detected object's visual appearance, temporal persistence, and spatial dynamics to estimate its semantic importance; an iterative, prompt-driven refinement loop then repeatedly compares the generated caption against high-prominence objects and injects contextually relevant, missing entities. Across MSVD and MSR-VTT~\cite{xu2016msr}, the resulting captions achieve consistently higher object-grounded completeness and lower inconsistency than both the base captioner and single-pass prompting, a finding corroborated by human evaluation and qualitative comparison against ChatGPT and Gemini.

The primary contributions of this work are as follows:
\begin{itemize}
    \item \textbf{Prominence-aware semantic modeling:} We introduce a novel prominence formulation that jointly models object appearance, temporal persistence, and spatial dynamics to estimate the semantic importance of objects within a video.
    \item \textbf{Post-hoc, iterative prominence-guided refinement:} We propose a lightweight, model-agnostic framework that iteratively incorporates missing high-prominence objects into the caption through structured, LLM-based prompting, without retraining the underlying captioning model.
    \item \textbf{Comprehensive evaluation:} We validate the proposed approach through automatic object-grounded metrics, human evaluation, and qualitative comparison with state-of-the-art LLMs such as ChatGPT and Gemini.
\end{itemize}
Overall, the proposed framework offers a lightweight, scalable, and model-agnostic strategy for improving the faithfulness and completeness of video captioning systems.

The remainder of the paper is organized as follows: Section~\ref{sec:related_work} reviews related literature; Section~\ref{sec:methodology} details the proposed architecture; Section~\ref{sec:experimental_setup} describes the datasets, metrics, and experimental protocol; Section~\ref{sec:results} presents the quantitative, human, and qualitative results; and Section~\ref{sec:conclusion} concludes with directions for future work.
\section*{Code and Data Availability}
The source code implementing the proposed ProCap framework including the prominence scoring, rectification, and iterative refinement modules is publicly available at: \url{https://github.com/Debjyoti-Adhikary/ProCap}. The MSR-VTT and MSVD benchmark datasets used for evaluation are publicly available from their original sources.

\section{Related Work}
\label{sec:related_work}

\subsection{Classical Encoder Decoder Video Captioning}
\label{sec:rw_classical}
The Classical video captioning models were built on the idea of sequence-to-sequence frameworks. Early approaches used Convolutional Neural Networks (CNNs) to encode the visual frames from the input video and Recurrent Neural Networks (typically LSTMs) to generate the final descriptions \cite{naik2024video, gao2017video}. These models often incorporated temporal attention to focus on salient frames~\cite{yan2019stat}. For example,~\cite{vinyals2015show} introduced a CNN–LSTM model that achieved good BLEU/CIDEr scores on benchmarks. At this stage, success was measured largely by n-gram overlap metrics (BLEU, METEOR, CIDEr), so research emphasized improving these scores through encoder–decoder architectures. Because this approach of generating captions achieved high metric scores, it became the standard baseline in video captioning \cite{pan2016hierarchical,zhang2020dense}. Yet such models depended heavily on matching reference captions, often at the expense of deeper semantic understanding.

\subsection{Grounding Based and Saliency Aware Captioning}
\label{sec:rw_grounding}
To address faithfulness, grounding-based and object-aware methods were proposed. These approaches integrate explicit visual detectors or attention over object regions to ground the captions in image content~\cite{ma2019salient,kong2021self}. For example,~\cite{zhang2023learning} and \cite{wang2022m3s} incorporate object detectors or scene graphs so that the generated words align with detected entities. These models improve on hallucination by filtering generated captions against detected objects: if an object appears in the frame but is not mentioned, the model can be guided to include it. Similarly, some methods use visual grounding objectives or auxiliary classification loss to ensure objects are described. Because grounding ties words to actual pixels, these works reduced obvious errors. However, they have limitations. Most grounding approaches still rely on end-to-end training and do not guarantee inclusion of all important objects. They often treat all detections equally, lacking a mechanism to prioritize prominent entities. In other words, grounding methods improved consistency but did not systematically ensure completeness, and they often required retraining the model on combined tasks (captioning + object detection).

\subsection{Post Hoc Rectification Frameworks}
\label{sec:rw_rectification}
Recently, post-hoc rectification frameworks have been introduced to correct captioning errors without retraining the base model. In image captioning,~\cite{das2024reframe} proposed ReFrame, which plugs an object detector on top of a pre-trained captioner. ReFrame inspects the initial caption and adds missing objects (or removes hallucinated ones) to improve consistency. Inspired by this, video captioning work like ReCap~\cite{adhikary2025recap} applies a two-stage pipeline: it first generates captions for key frames, then uses a detector to filter and refine them before summarizing into a final caption. These rectification methods explicitly enforce that only detected objects appear in the text. Because they operate after caption generation, they can be model-agnostic plug-ins. Nevertheless, they share certain shortcomings. Most existing rectifiers perform a single pass over the caption and do not distinguish which objects are more important. In other words, they ensure correctness but not exhaustiveness. Moreover, they are not iterative: once corrected, the caption is final. Thus, rectification approaches improved factual accuracy but still lacked a mechanism to add missing content in multiple rounds.

\subsection{Large Vision Language Models and Prompt-Based Refinement}
\label{sec:rw_llm}
Large Vision Language Models (VLMs) and prompting have brought new strategies for refinement. Some recent works use powerful LLMs (e.g. GPT-based models) by feeding them frame features or transcripts, effectively summarizing videos with minimal task-specific design~\cite{tang2025video,brimont2026survey}. These LLM-based pipelines generate fluent descriptions, but they can still miss details unless specifically guided. Prompt-based methods have been used to improve captions by asking the model to include overlooked objects or verify content. For example, one can prompt an LLM with detected object names and instruct it to rewrite the caption to include them. These methods leverage general language understanding but typically require carefully crafted prompts and do not inherently rank object importance. As a result, LLM based captioning models share some gaps: they excel at language but need explicit visual cues to ensure completeness.

\subsection{Limitations of Reference-Based Evaluation Metrics}
\label{sec:rw_metrics}
Reliance on n-gram metrics has limitations. BLEU and similar scores reward phrasing that matches training captions but do not ensure the caption faithfully describes the video content~\cite{ullah2022thinking,liu2023models}. For example, a model can achieve high BLEU by using common phrases (e.g. "a person is doing something"), even if it omits or hallucinates details. In practice, classical models frequently hallucinate objects not present in the video or miss relevant elements (e.g. failing to mention a dog that appears). This discrepancy between metric scores and factual accuracy has been noted in recent analysis work~\cite{guan2024hallusionbench}. Because these pure generation models lack explicit grounding, high automatic scores did not prevent mistakes in object coverage. Hence, researchers observed that generation-based methods were insufficiently faithful: they often failed to capture all salient visual content.

\subsection{Research Gaps and Challenges}
\label{sec:rw_gaps}
Despite significant advancements in video captioning, existing approaches continue to suffer from limitations related to semantic grounding, object hallucination, and incomplete coverage of salient visual entities. Classical encoder-decoder architectures (Section~\ref{sec:rw_classical}) often generate generic captions with weak visual alignment, while recent LLM-based methods (Section~\ref{sec:rw_llm}) improve fluency at the cost of introducing externally inferred knowledge and hallucinated content. Although rectification-based frameworks such as ReFrame and ReCap (Section~\ref{sec:rw_rectification}) improve consistency through object-aware correction, they primarily focus on frame-level grounding and do not address missing semantic content at the video-caption level.

In summary, the literature shows a clear progression: generation-only models were accurate by metrics but ungrounded, grounding-based models reduced hallucination but still missed objects, and rectification approaches corrected errors post-hoc but in a one-shot manner. Because purely generative methods failed to be faithful, researchers proposed grounding and rectification frameworks. But those remedies still lacked mechanisms for emphasizing prominent objects and iteratively refining captions. Compounding this, as discussed in Section~\ref{sec:rw_metrics}, the reference-based metrics used to evaluate these methods are themselves ill-suited to expose such omissions, leaving both the modeling and measurement of semantic completeness as open problems. Hence, our work introduces a prominence-aware iterative rectification framework -- detailed in Section~\ref{sec:methodology} -- that goes beyond single-pass correction by ranking missing objects and re-generating captions in multiple rounds, leading to more complete and faithful video descriptions.

\section{Proposed Architecture of ProCap} \label{sec:methodology}
This section details the proposed architecture illustrated in Fig.~\ref{fig:teaser_figure}. The functionality of each module is presented individually as explained in the Algorithm~\ref{alg:overall}, after which their interaction and integration into the complete framework are discussed in the final subsection.
\begin{algorithm}[t]
\caption{Prominence-Guided Iterative Video Captioning Rectification}
\label{alg:overall}
\begin{algorithmic}[1]
\Require Video $\mathcal{V}$; max refinement iterations $R$; prominence threshold $\tau$
\Ensure Caption variants $y_{\text{short}}$, $y_{\text{long}}$, $\hat{y}^{(r)}$ (refined)

\Statex \textit{// Frame-level captioning and rectification (Sec.~\ref{sec:rectify})}
\State $\{f_1, \dots, f_T\} \gets$ \Call{ExtractKeyFrames}{$\mathcal{V}$}
\State $\mathcal{S} \gets \emptyset$, \quad $\mathcal{O} \gets \emptyset$
\For{each key frame $f_t$}
    \State $s_t \gets$ \Call{RecursiveCaption}{$f_t$} \Comment{mPLUG}
    \State $\mathcal{D}_t \gets$ \Call{Detect}{$f_t$} \Comment{YOLOv8 rectifier}
    \State $s_t \gets$ \Call{Rectify}{$s_t, \mathcal{D}_t$} \Comment{prune ungrounded tokens}
    \State $\mathcal{S} \gets \mathcal{S} \cup \{s_t\}$; \quad $\mathcal{O} \gets \mathcal{O} \cup \mathcal{D}_t$
\EndFor

\Statex \textit{// Prominence-aware object scoring (Sec.~\ref{sec:prominence})}
\For{each object $o_i \in \mathcal{O}$}
    \State compute $A_i$, $P_i$, $D_i$ \Comment{Defs.~1--3}
    \State $\Pi(o_i) \gets A_i \cdot (P_i + D_i - P_i D_i)$ \Comment{Def.~4}
\EndFor
\State $\mathcal{O}^* \gets \{o_i \in \mathcal{O} \mid \Pi(o_i) \geq \tau\}$, ranked by $\Pi(o_i)$ \label{ln:filter}

\Statex \textit{// Single-pass prompt variants (Sec.~\ref{sec:promptvariants})}
\State $y_{\text{short}} \gets \mathcal{F}_{\text{LLM}}(\mathcal{S}, \mathcal{O}^*; \text{prompt}_{\text{short}})$
\State $y_{\text{long}} \gets \mathcal{F}_{\text{LLM}}(\mathcal{S}, \mathcal{O}^*; \text{prompt}_{\text{long}})$

\Statex \textit{// Iterative refinement loop (Sec.~\ref{sec:refine})}
\State $\hat{y}^{(0)} \gets \mathcal{F}_{\text{LLM}}(\mathcal{S})$ \label{ln:init}
\For{$r = 1$ \textbf{to} $R$}
    \State $\mathcal{M}^{(r)} \gets \{o \in \mathcal{O}^* \mid o \notin \hat{y}^{(r-1)}\}$ \label{ln:missingobj}
    \If{$\mathcal{M}^{(r)} = \emptyset$}
        \State \textbf{break} \Comment{no remaining semantic gap}
    \Else
        \State $\hat{y}^{(r)} \gets \mathcal{F}_{\text{LLM}}(\hat{y}^{(r-1)}, \mathcal{M}^{(r)}, \mathcal{S})$ \label{ln:refineupdate}
    \EndIf
\EndFor

\State \Return $y_{\text{short}}$, $y_{\text{long}}$, $\hat{y}^{(r)}$
\end{algorithmic}
\end{algorithm}
\subsection{Comprehensive Video Captioning via Rectification}~\label{sec:rectify}
As established in Section~\ref{sec:rw_rectification}, ReCap~\cite{adhikary2025recap} provides the architectural foundation for this work: a two-stage pipeline that rectifies frame-level captions against an external object detector before summarizing them into a single video-level caption via a single LLM pass, treating all detected objects as equally important. The proposed framework, ProCap, retains ReCap's frame-level rectification stage unchanged (Fig.~\ref{fig:teaser_figure}, as shown in the example~\ref{ex:frame_captions}) while directly addressing both limitations: Section~\ref{sec:prominence} introduces prominence-guided object ranking, and Section~\ref{sec:refine} replaces the single summarization pass with an iterative refinement loop.

\begin{example}[Frame-Level Caption Inconsistency Across a Video]
\noindent
\label{ex:frame_captions}
Tix frames sampled extracted from the key frames from a single video depicting a mother and
her baby. A frame-level captioning model generated as per~\cite{adhikary2025recap} produces the following captions independently for each frame:
\begin{small}
\begin{enumerate}
    \item \textit{``baby boy in a cradle''}
    \item \textit{``person is a mother of two''}
    \item \textit{``person in the hospital with her baby''}
    \item \textit{``person in the hospital with her newborn baby''}
    \item \textit{``a woman holding a baby in her arms while sitting
    at a table''}
    \item \textit{``a woman and a child in a kitchen with a table''}
\end{enumerate}
\end{small}
\end{example}




\subsection{Prominence Aware Object Scoring}~\label{sec:prominence}

Once we obtain the processed captions for all frames capturing the significant events of the video, we employ an object segmentation model (YOLOv8, consistent with the rectifier module) to extract the set of objects present across the video~\cite{varghese2024yolov8}. In order to preserve the temporal alignment between detected objects and the corresponding frame-level captions, and to systematically quantify object saliency, spatial significance, and motion characteristics, we introduce a prominence-based scoring mechanism. This mechanism serves as the theoretical bridge between raw, unstructured object proposals and narratively cohesive semantic selection. Formulating our proposed prominence metric requires abstracting the continuous, complex spatio-temporal elements of a video into a tractable scalar value that accurately reflects human perceptual attention. In the evaluated rectification framework, the prominence of an object instance, denoted as $\Pi(o_i)$, is mathematically decomposed into three distinct, normalized components: Appearance ($A_i$), Presence ($P_i$) and Dynamics ($D_i$). These components are subsequently fused using a probabilistic union-inspired formulation.

Let $\mathcal{O} = \{o_1, o_2, \dots, o_N\}$ denote the set of detected object instances across the video consisting of $T$ frames. For each object $o_i$, we define its prominence score $\Pi(o_i)$ as a function of three normalized components: Appearance ($A_i$), Presence ($P_i$), and Dynamics ($D_i$), formalized below.

The appearance component captures the spatial importance of an object within a frame.

\begin{definition}[Appearance]
\label{def:appearance}
The appearance score of object $o_i$ is defined as the ratio of the object's bounding box area to the total frame area, averaged over all frames in which the object is detected:
\[
A_i = \frac{1}{|F_i|} \sum_{t \in F_i} \frac{\text{area}(o_i^t)}{\text{area}(\text{frame})}
\]
where $F_i \subseteq \{1, \dots, T\}$ denotes the set of frames in which object $o_i$ is detected.
\end{definition}

This ensures that objects occupying a larger visual region—hence more likely to be perceptually significant—are assigned higher importance.

The presence component models the temporal persistence of an object, capturing the intuition that objects consistently present throughout the video are more semantically relevant.

\begin{definition}[Presence]
\label{def:presence}
The presence score of object $o_i$ is defined as the fraction of frames in which the object appears relative to the total number of frames in the video:
\[
P_i = \frac{|F_i|}{T}
\]
\end{definition}

Additionally, we incorporate Dynamics to model the interaction and motion characteristics of an object.

\begin{definition}[Dynamics]
\label{def:dynamics}
The dynamics score of object $o_i$ is defined as the normalized average change in spatial distance between the object and other co-occurring objects across frames:
\[
D_i = \frac{1}{|F_i|} \sum_{t \in F_i} \left( \frac{1}{|\mathcal{O}_t| - 1} \sum_{o_j \in \mathcal{O}_t, j \neq i} \frac{\|c_i^t - c_j^t\|_2}{d_{\max}} \right)
\]
\end{definition}

This formulation captures relational movement and interaction intensity, thereby emphasizing objects that play an active role in the scene rather than remaining static or peripheral.

To combine these factors into a single measure of object importance, we define the prominence score using a \textbf{probabilistic union-inspired} formulation:

\begin{definition}[Prominence Score]
\label{def:prominence}
Given the Appearance, Presence, and Dynamics components of object $o_i$, the prominence score $\Pi(o_i)$ is defined as:
\[
\Pi(o_i) = A_i \cdot (P_i + D_i - P_i D_i)
\]
\end{definition}

This formulation can be interpreted as modeling the joint contribution of temporal persistence and dynamic interaction using an inclusion--exclusion principle, where $(P_i + D_i - P_i D_i)$ represents the score of an object being either persistently present or dynamically significant. The multiplicative interaction with $A_i$ ensures that only visually salient objects are emphasized, thereby preventing small or insignificant objects from dominating the ranking despite high temporal or dynamic scores.

\begin{figure}[!t]
    \centering
    \subfloat[Frame 1]{\includegraphics[width=0.48\columnwidth]{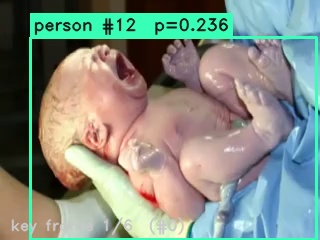}}\hfill
    \subfloat[Frame 2]{\includegraphics[width=0.48\columnwidth]{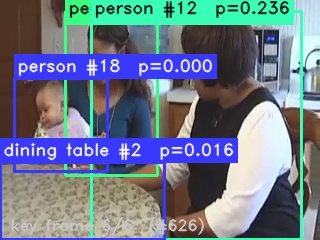}}
    \caption{Visualization of object prominence estimation across three representative frames.}
    \label{fig:prominence_example}
\end{figure}

To further illustrate the effectiveness of the proposed prominence formulation, Fig.~\ref{fig:prominence_example} presents an example visualization of object prominence estimation across three representative frames sampled from an input video. Each frame contains the detected objects annotated with bounding boxes produced by the rectifier module. The visualization demonstrates how the proposed framework jointly considers spatial saliency, temporal persistence, and relational dynamics to determine the semantic importance of objects within the video.

\begin{example}[Prominence Scoring from the Example~\ref{ex:frame_captions}]
\noindent
\label{ex:prominence}
For the same video as Fig.~\ref{fig:teaser_figure} and
\ref{fig:prominence_example}, the detected \textit{person} (the baby)
receives a substantially higher prominence score than the background
\textit{dining table}:
\begin{small}
\begin{align*}
\textit{person:}\quad
& A=0.59,\;
P=0.40,\;
D=0.0004,\;
\mathbf{\Pi=0.2361} \\
\textit{dining table:}\quad
& A=0.1676,\;
P=0.0973,\;
D=0.0004,\;
\mathbf{\Pi=0.0164}
\end{align*}
\end{small}
Despite both objects having negligible dynamics ($D\approx0.0004$), the
person's prominence is approximately \textbf{14$\times$} that of the
dining table, primarily due to its larger appearance ratio and longer
temporal persistence.
\end{example}


These values indicate that the \textit{person} occupies a substantially larger spatial region within the frame sequence and persists for a longer temporal duration, resulting in a considerably higher prominence score. In contrast, although the \textit{dining table} is consistently detected, its relatively smaller appearance ratio and lower persistence reduce its overall semantic significance. The low dynamics values in both cases further suggest limited inter-object motion, indicating that prominence in this example is primarily governed by appearance and temporal presence.

The figure also highlights the temporal continuity of object tracking through the associated timestamps and centroid locations. The detected \textit{person} remains visible between timestamps \textbf{00:15} and \textbf{00:24}, whereas the \textit{dining table} is observed only between \textbf{00:17} and \textbf{00:22}. This temporal information contributes directly to the persistence component of the prominence score, enabling the framework to distinguish primary scene entities from secondary or background objects.

The resulting prominence score $\Pi(o_i)$ thus provides a principled measure of object importance by jointly capturing spatial saliency, temporal consistency, and relational dynamics. These scores are subsequently used to rank detected objects and guide the iterative caption refinement process, enabling the system to prioritize semantically significant entities that are more likely to contribute to a complete and faithful video description.

\subsection{Iterative Refinement Loop}
\label{sec:refine}
Although a single-pass caption generation mechanism can produce coherent summaries from frame-level descriptions, it often fails to capture all semantically relevant objects, leading to incomplete or partially grounded captions. We therefore introduce an iterative refinement loop -- formalized as the closing stage of Algorithm~\ref{alg:overall} -- that progressively improves caption quality by explicitly identifying and incorporating missing yet prominent objects, treating caption generation as a feedback-driven process in which successive iterations correct omissions without retraining the underlying model.

Given a video $\mathcal{V}$, the pipeline first aggregates frame-level captions $\mathcal{S}$ and extracts a set of prominent objects $\mathcal{O}^*$ using the prominence-aware scoring mechanism (Algorithm~\ref{alg:overall}, line~\ref{ln:filter}). The iterative refinement process then generates a sequence of captions $\{\hat{y}^{(0)}, \hat{y}^{(1)}, \dots, \hat{y}^{(R)}\}$, where $\hat{y}^{(0)}$ is the initial caption and each subsequent iteration refines it by incorporating missing semantic elements.

\textbf{Initial caption (iteration 0).} A base caption is generated solely from the aggregated frame-level descriptions (Algorithm~\ref{alg:overall}, line~\ref{ln:init}):
\[
\hat{y}^{(0)} = \mathcal{F}_{\text{LLM}}(\mathcal{S})
\]
where $\mathcal{F}_{\text{LLM}}$ denotes the prompt-conditioned generation function of the language model. The prompt enforces conciseness (typically 10--12 words) and instructs the model to remove redundant or contextually inconsistent frame-level statements, producing a fluent but potentially incomplete caption that initializes refinement.

\textbf{Missing-object identification.} For each subsequent iteration ($r > 0$), we identify objects that are semantically relevant but absent from the current caption via a lexical grounding check (Algorithm~\ref{alg:overall}, line~\ref{ln:missingobj}):
\[
\mathcal{M}^{(r)} = \{o \in \mathcal{O}^* \mid o \notin \hat{y}^{(r-1)}\}
\]
capturing the semantic gap between visual evidence and textual description.

\textbf{Refinement.} If $\mathcal{M}^{(r)}$ is non-empty, the caption is updated as (Algorithm~\ref{alg:overall}, line~\ref{ln:refineupdate}):
\[
\hat{y}^{(r)} = \mathcal{F}_{\text{LLM}}(\hat{y}^{(r-1)}, \mathcal{M}^{(r)}, \mathcal{S})
\]
conditioning the model jointly on the current caption, the missing objects, and the original frame-level descriptions. This encourages the model to incorporate missing objects only when contextually consistent with the video content, while constraints on brevity and coherence prevent verbose or noisy outputs. The process is progressive: each iteration recomputes $\mathcal{M}^{(r)}$ against the previous output $\hat{y}^{(r-1)}$ and replaces it, preserving already-correct semantic elements while selectively enriching the caption, until either the semantic gap closes or a fixed number of iterations $R$ is reached. The following example shows the output of one video after the iterations.

\begin{example}[Iterative Refinement using the objects from Example~\ref{ex:prominence}]
\label{ex:refinement}

This example illustrates the refinement process for the video shown in
Fig.~\ref{fig:teaser_figure} and Fig.~\ref{fig:prominence_example}.

\begin{itemize}
    \item \textbf{Iteration 0:} ``A newborn baby is seen inside a hospital room.''

    \item \textbf{Iteration 1 (+ mother):} ``A mother cradles her newborn baby inside a hospital room.''

    \item \textbf{Iteration 2 (Final + dining table):}
    ``A mother cradles her newborn baby while sitting at a table in the hospital.''
\end{itemize}
\end{example}
(See Appendix~D for the complete pipeline output, including the single-pass Short/Long variants of Section~\ref{sec:promptvariants}.)

\subsection{Designing the Prompt}\label{sec:promptvariants}
To transform frame-level captions into a coherent, semantically faithful video-level description, we design a structured prompt engineering strategy that leverages large language models as a lightweight, training-free post-processing module. Let $\mathcal{S} = \{s_1, s_2, \dots, s_K\}$ denote the set of frame-level captions and $\mathcal{O} = \{o_1, o_2, \dots, o_M\}$ the set of detected objects derived from the prominence-aware module; the objective is to generate a consolidated caption $\hat{y}$ that maximizes semantic consistency while minimizing redundancy and hallucination:
\[
\hat{y} = \mathcal{F}_{\text{LLM}}(\mathcal{S}, \mathcal{O}; \theta)
\]
where $\mathcal{F}_{\text{LLM}}$ is the prompt-conditioned generation function of the language model, parameterized by decoding hyperparameters $\theta$ and instantiated through the four concrete templates in Table~\ref{tab:prompt_templates}: a short, a medium-length, and a long descriptive variant, plus an object-insertion clause conditioning generation on $\mathcal{O}$.
\begin{table}[!t]
\centering
\caption{Prompt templates used for caption generation.}
\label{tab:prompt_templates}
\footnotesize
\begin{tabularx}{\columnwidth}{@{}p{0.22\columnwidth}X@{}}
\toprule
\textbf{Prompt Type} & \textbf{Template} \\
\midrule
Short &
\texttt{Combine the following sentences from frames of a video into a single concise caption: \{Frame captions\}. Remove sentences that do not match the overall context.} \\
\addlinespace
Medium &
\texttt{Combine the following sentences from frames of a video into a single 10--12 word caption. Remove sentences that do not match the overall context: \{Frame captions\}. Output only the caption.} \\
\addlinespace
Long &
\texttt{Combine the following sentences from frames of a video into a single detailed 15--20 word caption. Remove sentences that do not match the overall context: \{Frame captions\}. Output only the caption.} \\
\addlinespace
Object Insertion &
\texttt{Additionally, if relevant, include these objects: }
$\mathcal{O}=\{o_1,o_2,\ldots,o_M\}$. \\
\bottomrule
\end{tabularx}
\end{table}
These prompt variants impose explicit lexical and structural constraints on the generated output, effectively controlling the trade-off between brevity and semantic richness. Beyond these surface-level templates, every variant is conditioned on a common instruction schema that enforces four consistent design constraints on the language model, regardless of which prompt type is selected:

\begin{center}
\noindent
\fbox{
\begin{minipage}{0.92\columnwidth}
\centering \textit{Shared Instruction Schema}
\begin{enumerate}
    \item Identify non-redundant core events
    \item Generate a compact, fluent summary (1--2 sentences)
    \item Include only objects present in the captions or object list
    \item Avoid hallucination or external inference
\end{enumerate}
\end{minipage}
}
\end{center}

Together, the template-level constraints in Table~\ref{tab:prompt_templates} and this shared instruction schema ensure that every generated caption, irrespective of length, remains concise, non-redundant, and strictly grounded in the aggregated frame captions $\mathcal{S}$ and prominent object set $\mathcal{O}$, forming the initialization point for the iterative refinement process described next.

\subsection{System Integration}
\begin{figure}[!t]
    \centering
    \includegraphics[width=\columnwidth]{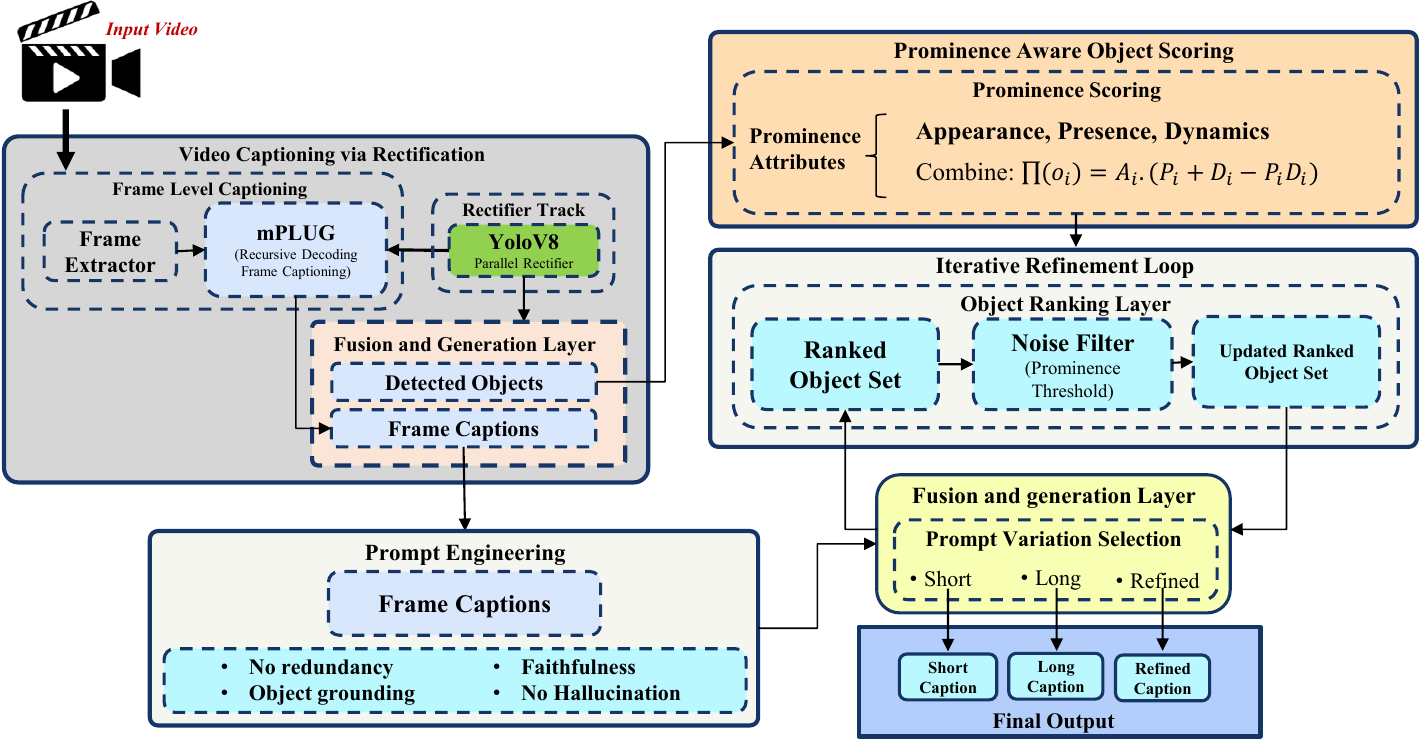}
    \caption{Proposed Architecture illustrating the working and flow of constituent modules.}
    \label{new_architecture}
\end{figure}

Fig.~\ref{new_architecture} and Algorithm~\ref{alg:overall} jointly summarize how the preceding modules: frame-level rectification (Section~\ref{sec:rectify}), prominence scoring Section~\ref{sec:prominence}), and iterative refinement (Sections~\ref{sec:refine}, \ref{sec:promptvariants}) compose into the unified ProCap pipeline: each stage's output (rectified frame captions, ranked objects, structured prompts) feeds directly into the next, operating end-to-end as a modular, plug-and-play layer over the frozen captioning backbone with no retraining required.
The aggregated frame captions and selected prominent objects are subsequently encoded into structured prompts and passed to the language model to generate an initial video-level caption. This caption is then iteratively refined by identifying missing prominent objects and updating the prompt accordingly, progressively improving semantic completeness while preserving fluency. Overall, the framework functions as a modular, plug-and-play enhancement layer that operates on top of existing video captioning models, requiring no retraining while enabling scalable and efficient caption refinement.

\section{Implementation Details and Setup}\label{sec:experimental_setup}

\subsection{Implementation Details}
\label{sec:implementation}
All stages of the proposed pipeline, frame level captioning, YOLOv8-based rectification, prominence scoring, and LLM-based prompt refinement were executed on a dedicated GPU compute cluster of 4 nodes, each with dual 12-core/24-thread Intel Xeon (Skylake) processors, two NVIDIA Tesla P100 GPUs (16\,GB HBM2, 3584 CUDA cores each), and 128\,GB RAM, running Ubuntu 22.04.5 LTS. The underlying captioning backbone, mPLUG, was not trained from scratch: consistent with the previously established ReCap framework~\cite{adhikary2025recap}, it was trained following the protocol and hyperparameters of its original publication~\cite{xu2023mplug}, and the resulting pretrained checkpoint is reused unchanged -- the entire prominence-guided rectification and refinement pipeline operates on top of this fixed, frozen backbone without any additional fine-tuning or retraining.

\subsection{Datasets}
We evaluate the proposed framework on two widely adopted benchmarks for
single-sentence video captioning: \textbf{MSR-VTT}
and \textbf{MSVD}~\cite{xu2016msr}. Both are standard testbeds in recent
literature, allowing us to assess generalization across videos of varying
complexity, scene diversity, and object distributions.

\subsubsection{MSR-VTT}
MSR-VTT is one of the largest open-domain video captioning benchmarks,
comprising \textbf{10,000} web video clips spanning \textbf{20 semantic
categories} (e.g., sports, music, cooking, news, gaming). Each
\textbf{10--30 second} clip is paired with about \textbf{20 human-written
captions}, yielding nearly \textbf{200,000} annotations. We follow the
standard split of \textbf{6,513 training}, \textbf{497 validation}, and
\textbf{2,990 testing} videos.

\subsubsection{MSVD}
MSVD (YouTube2Text) contains \textbf{1,970} YouTube clips of everyday
human activities, each roughly \textbf{10 seconds} long and annotated
with nearly \textbf{40 English descriptions} (about \textbf{80,000}
captions total). We use the standard split of \textbf{1,200 training},
\textbf{100 validation}, and \textbf{670 testing} videos.

\subsection{Human Evaluation via Survey-Based Assessment}\label{sec:human_eval}
To complement standard automatic evaluation metrics and better capture the qualitative aspects of caption quality (see Section~\ref{sec:eval_metrics} for why lexical-overlap metrics fall short), we conducted a structured, human-centered evaluation study.

\subsubsection{Study Design and Participants}
The survey was administered through a custom-built web interface accessible via a unique link, completed self-paced by each participant. Each task page presented a single video alongside the five caption variants (Table~\ref{tab:caption_variants}), displayed in a randomized, vertically stacked layout under blinded labels (\emph{Caption 1--5}) rather than method names, to prevent bias toward recognizable systems; both video order and caption order were independently randomized per participant. A total of 100 clips were selected via stratified random sampling over video duration, drawn from a pool pre-filtered to exclude degenerate clips offering insufficient visual grounding for comparison; this sampling procedure was pre-registered before annotation to prevent post-hoc selection of favorable examples. A total of 110 proficient English speakers participated, with no Computer Vision or NLP background required; the study followed the ethical guidelines of the Indian Institute of Technology Kharagpur, with informed, voluntary, and anonymized participation.

Participants rated each caption along two dimensions on a five-point Likert scale ($1=$ lowest, $5=$ highest, no intermediate anchors), judging only visible video content and rating each dimension independently of the other (see Appendix~A for a representative screenshot of the deployed interface).:
\begin{itemize}
    \item \textbf{Completeness (higher is better):} how well the caption covers all important objects, people, and events visible in the video.
    \item \textbf{Inconsistency (lower is better):} the degree to which the caption contains information that is factually incorrect, hallucinated, or unsupported by the video.
\end{itemize}

\subsubsection{Caption Variants and Score Aggregation}
For each video, annotators rated all five variants shown in Table~\ref{tab:caption_variants}: the benchmark ground truth, the mPLUG-2 baseline, the short and long prompt-based variants, and the final iteratively refined caption. Ground truth serves as an upper-bound completeness reference rather than a perfect-inconsistency baseline, since benchmark annotations can themselves contain minor inaccuracies; mPLUG-2 serves as the primary external baseline.

\begin{table}[h]
\centering
\caption{Description of caption variants evaluated per video.}
\label{tab:caption_variants}
\begin{tabular}{lp{0.58\linewidth}}
\textbf{Label shown to annotator} & \textbf{Description} \\
Caption 1-5 (randomized) & Ground-truth annotation from the benchmark dataset \\
                         & mPLUG-2 output (state-of-the-art baseline) \\
                         & Short prompt-based variant (concise, $\leq$12 words) \\
                         & Long prompt-based variant (detailed, 15-20 words) \\
                         & Final iteratively refined caption \\
\end{tabular}
\end{table}

Ratings were aggregated at two levels: for each video $v$ and variant $c$, scores were first averaged across all $N$ annotators,
\begin{equation}
\overline{S}_{v,c}=\frac{1}{N}\sum_{i=1}^{N}S_{v,c}^{(i)}
\end{equation}
then averaged again across all $|V|$ evaluated videos,
\begin{equation}
\overline{S}_{c}=\frac{1}{|V|}\sum_{v\in V}\overline{S}_{v,c}
\end{equation}
ensuring the reported figures reflect system-level performance rather than any single high- or low-difficulty instance. Percentage improvements reported in Section~4 are the relative change from the baseline group $b$ (base caption or mPLUG-2) to the proposed variant group $c$:
\begin{equation}
\Delta_{c+b}=\frac{\overline{S}_{c}-\overline{S}_{b}}{\overline{S}_{b}}\times100\%
\end{equation}
where a positive $\Delta$ is desirable for completeness and a negative $\Delta$ is desirable for inconsistency.

\subsection{Evaluation Metrics}\label{sec:eval_metrics}
Although the proposed framework introduces a prominence-guided iterative refinement strategy, the primary objective is inspired from ReCap~\cite{adhikary2025recap}, i.e., generating captions that are both semantically complete and visually faithful. Conventional captioning metrics such as BLEU, METEOR, ROUGE-L, CIDEr, and SPICE evaluate lexical similarity with reference captions and therefore fail to determine whether a generated caption accurately represents the visual content of a video. Since the proposed framework is model-agnostic and does not rely on ground-truth supervision, the generated captions may differ substantially from the reference annotations while still providing a more comprehensive and faithful description of the scene. Consequently, we adopt the object-grounded evaluation metrics proposed in ReCap, namely Temporal Completeness and Temporal Inconsistency, which directly quantify the semantic correspondence between the generated caption and the visually significant objects present in the video, formalized below.

\subsubsection{Temporal Completeness}
Temporal Completeness measures the proportion of significant objects present in the video that are successfully described in the generated caption, where an object is regarded as \emph{significant} if its temporal persistence exceeds a predefined threshold $\tau$, thereby filtering short-lived detections while emphasizing visually important entities.

\begin{definition}[Temporal Completeness]
\label{def:completeness}
Adapted from~\cite{adhikary2025recap}, given a generated caption $C_{\mathrm{cap}}$ for a video $V$ with $T$ frames and a set of significant objects $O = \{o_1, o_2, \ldots, o_n\}$, the Temporal Completeness score is defined as:
\begin{equation}
\label{eq:completeness}
C(C_{\mathrm{cap}}, O, V)
=
\frac{1}{|O|}
\sum_{i=1}^{|O|}
\mathbb{I}
\left(
o_i \in R(C_{\mathrm{cap}})
\land
\frac{|S(o_i)|}{T} \ge \tau
\right),
\end{equation}
where $R(C_{\mathrm{cap}})$ denotes the set of objects mentioned in $C_{\mathrm{cap}}$, $S(o_i)$ the temporal span of object $o_i$, and $\mathbb{I}(\cdot)$ the indicator function.
\end{definition}

A higher Temporal Completeness score indicates that the generated caption successfully captures a larger proportion of the visually significant content. Since the proposed prominence-guided refinement explicitly incorporates omitted but important objects into the caption, improvements in this metric directly reflect the effectiveness of the proposed refinement strategy.

\subsubsection{Temporal Inconsistency}
Temporal Inconsistency quantifies the complementary failure mode: the proportion of significant objects present in the video that are omitted from the generated caption, providing an object-centric estimate of semantic incompleteness and visual grounding errors.

\begin{definition}[Temporal Inconsistency]
\label{def:inconsistency}
Adapted from~\cite{adhikary2025recap}, using the same notation as Definition~\ref{def:completeness}, the Temporal Inconsistency score is defined as:
\begin{equation}
\label{eq:inconsistency}
I(C_{\mathrm{cap}}, O, V)
=
\frac{1}{|O|}
\sum_{i=1}^{|O|}
\mathbb{I}
\left(
o_i \notin R(C_{\mathrm{cap}})
\land
\frac{|S(o_i)|}{T} \ge \tau
\right).
\end{equation}
\end{definition}

Unlike conventional hallucination metrics that primarily penalize objects absent from the video, Temporal Inconsistency specifically captures missing salient objects, making it particularly suitable for evaluating post-hoc refinement frameworks whose primary objective is to improve semantic coverage rather than lexical similarity.

Within the proposed framework, the iterative refinement module repeatedly identifies high-prominence objects that remain absent from the current caption and incorporates them through structured prompting. Consequently, each refinement iteration is expected to reduce Temporal Inconsistency while simultaneously increasing Temporal Completeness.

\section{Experimental Results} \label{sec:results} 
In this section, we evaluate the effectiveness of the proposed prominence-guided iterative rectification framework through a combination of automatic metrics, human evaluation, and qualitative analysis. The results are structured to highlight (i) improvements in semantic completeness and reduction in inconsistency, (ii) validation through human-centered survey analysis, and (iii) illustrative qualitative examples demonstrating the impact of the proposed approach across datasets.

\subsection{Improvement in Completeness and Inconsistency}
We evaluate caption quality using the object-grounded Temporal Completeness and Temporal Inconsistency metrics defined in Eqs.~(4)--(5) (Section~\ref{sec:eval_metrics}), which respectively measure the coverage and omission of temporally-persistent, salient objects in the generated caption. The prominence-aware object selection mechanism is central to improving both: by prioritizing objects based on spatial saliency, temporal persistence, and relational dynamics, the framework directs refinement toward the entities most likely to be semantically significant, addressing the omissions observed in the base caption and mPLUG-2 baselines.

\begin{figure*}[!t]
\centering
\subfloat[MSRVTT: Completeness\label{fig:msrvtt_completeness}]
{\includegraphics[width=0.24\textwidth]{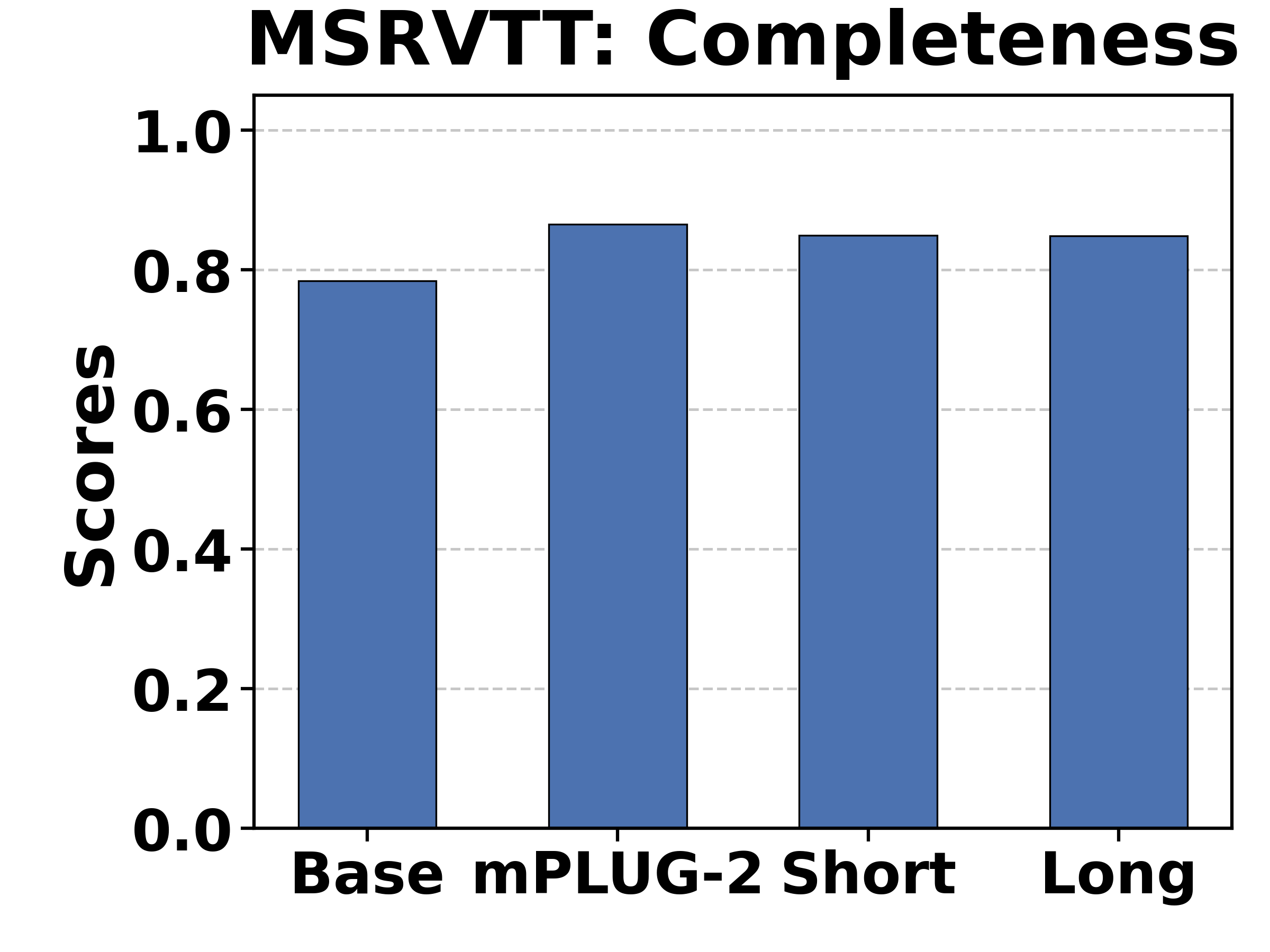}}
\hfill
\subfloat[MSRVTT: Inconsistency\label{fig:msrvtt_inconsistency}]
{\includegraphics[width=0.24\textwidth]{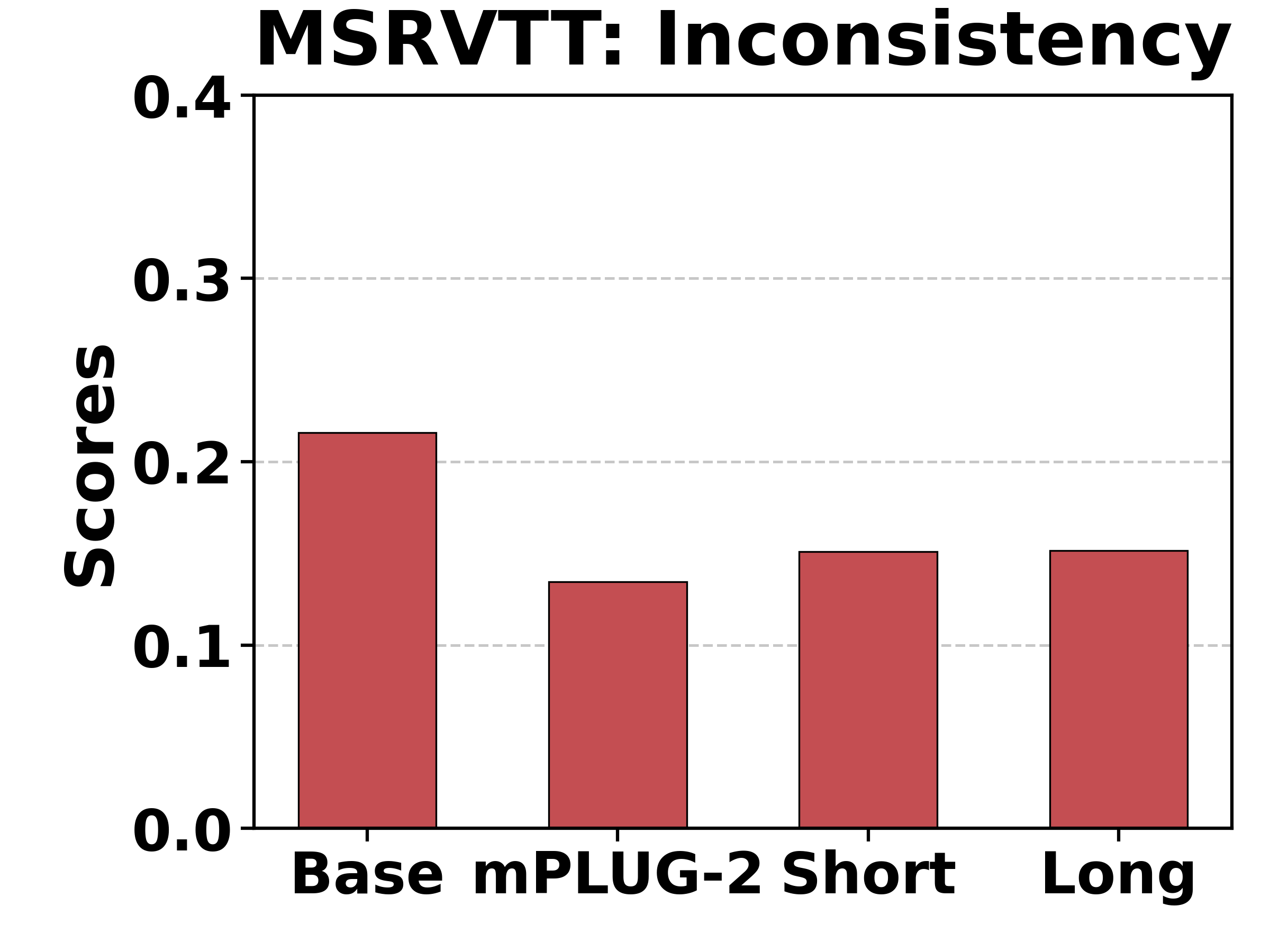}}
\hfill
\subfloat[MSVD: Completeness\label{fig:msvd_completeness}]
{\includegraphics[width=0.24\textwidth]{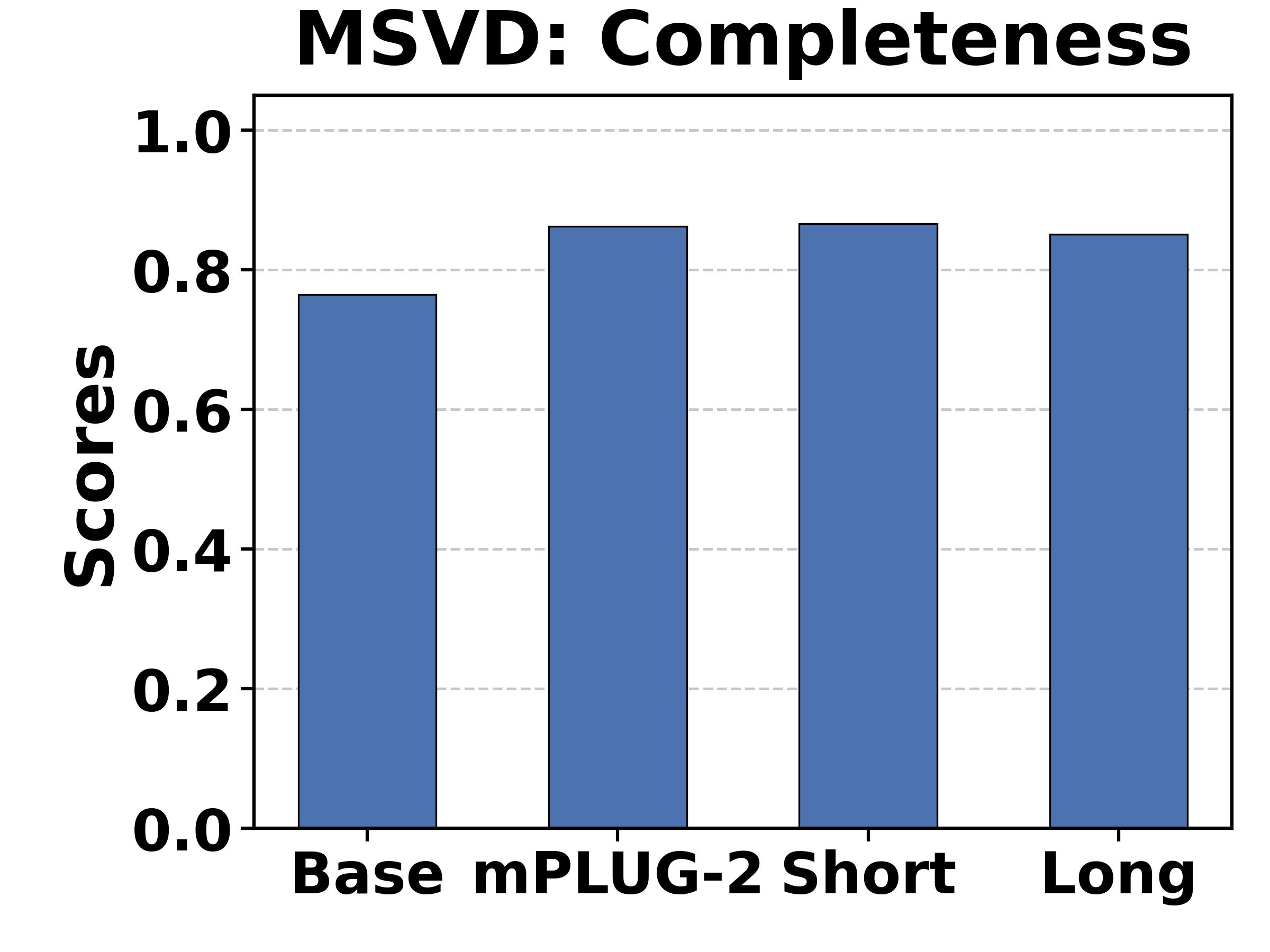}}
\hfill
\subfloat[MSVD: Inconsistency\label{fig:msvd_inconsistency}]
{\includegraphics[width=0.24\textwidth]{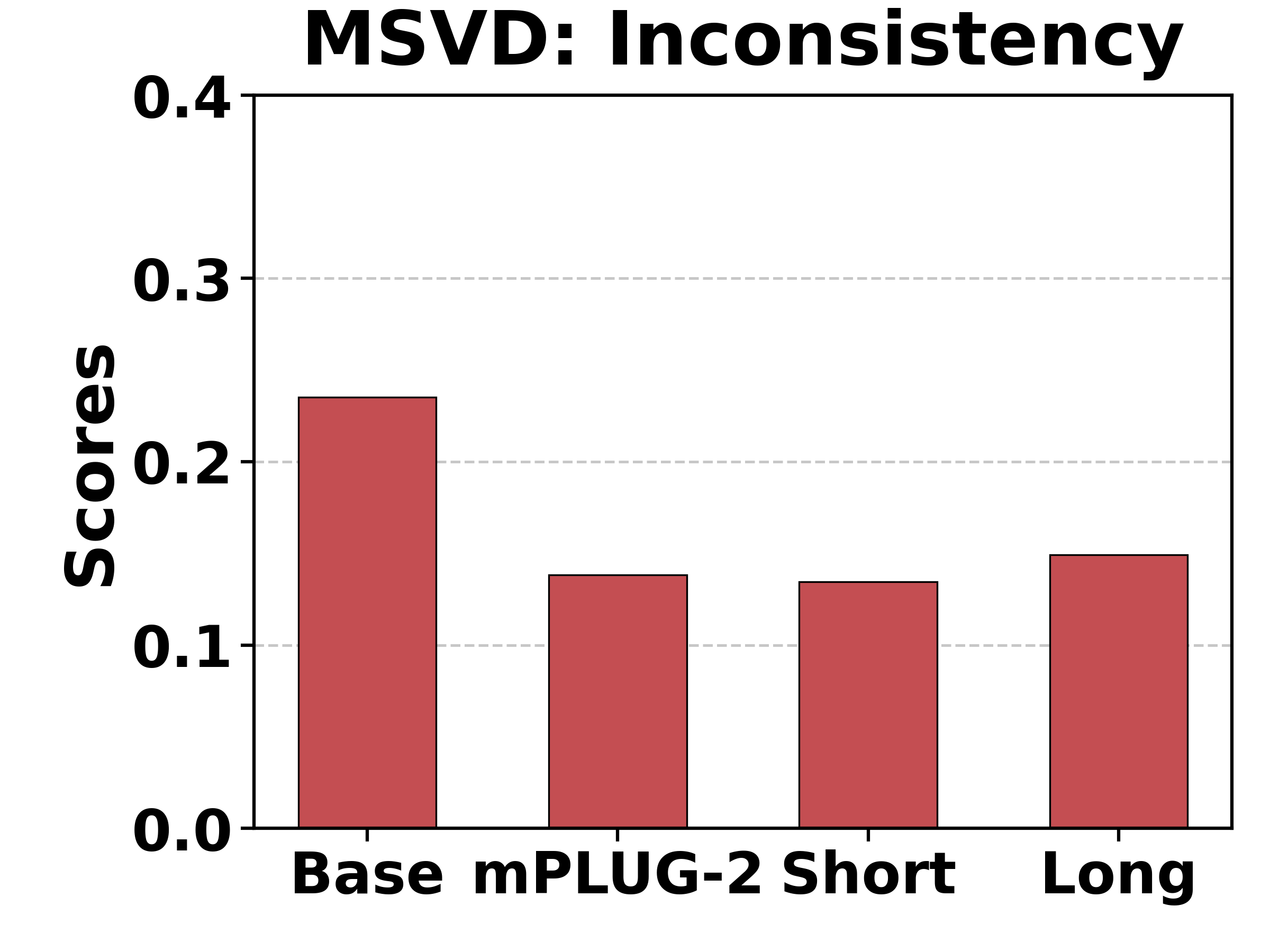}}
\caption{Automatic Temporal Completeness (higher is better) and Temporal Inconsistency (lower is better) on MSRVTT and MSVD, computed via Eqs.~(4)--(5). Bars are ordered: base (unrectified) caption, mPLUG-2 baseline, short-prompt variant, and long-prompt variant.}
\label{fig:comparison_metrics}
\end{figure*}

Table~\ref{tab:auto_delta} reports the resulting relative improvements (Eq.~(3)) for the short and long variants against both baselines, computed from Fig.~\ref{fig:comparison_metrics}. Both variants improve substantially over the unrectified base caption on both datasets, and the short variant on MSVD additionally edges out the mPLUG-2 baseline itself (\textbf{+0.43\%} completeness, \textbf{-2.75\%} inconsistency) despite mPLUG-2 already being a strong, fluent captioner. On the remaining variant-baseline pairs, the two variants are broadly comparable to mPLUG-2 on these automatic, coverage-based metrics, while both substantially outperform the unrectified base caption in every case. Section~\ref{human eval section} reports a clearer separation once caption quality is judged independently by human raters.

\begin{table}[t]
\centering
\caption{Relative improvement ($\Delta$, via Eq.~(3)) in automatic Temporal Completeness ($\Delta C$, higher is better) and Temporal Inconsistency ($\Delta I$, lower is better) for the short and long variants over each baseline.}
\label{tab:auto_delta}
\begin{tabular}{llcccc}
\hline
\textbf{Dataset} & \textbf{Variant} & \multicolumn{2}{c}{\textbf{vs.\ Base Caption}} & \multicolumn{2}{c}{\textbf{vs.\ mPLUG-2}} \\
 & & $\Delta C$ & $\Delta I$ & $\Delta C$ & $\Delta I$ \\
\hline
MSVD & Short & +13.19\% & -42.88\% & +0.43\% & -2.75\% \\
MSVD & Long & $\approx$+11\% & $\approx$-36\% & $\approx$-1\% & $\approx$+8\% \\
\hline
MSRVTT & Short & +8.29\% & -30.11\% & $\approx$-2\% & $\approx$+12\% \\
MSRVTT & Long & $\approx$+9\% & $\approx$-30\% & $\approx$-2\% & $\approx$+13\% \\
\hline
\end{tabular}
\vspace{2pt}
\end{table}

\subsection{Human Evaluation Results}\label{human eval section}
\begin{figure*}[!t]
    \centering
    \includegraphics[width=0.7\textwidth]{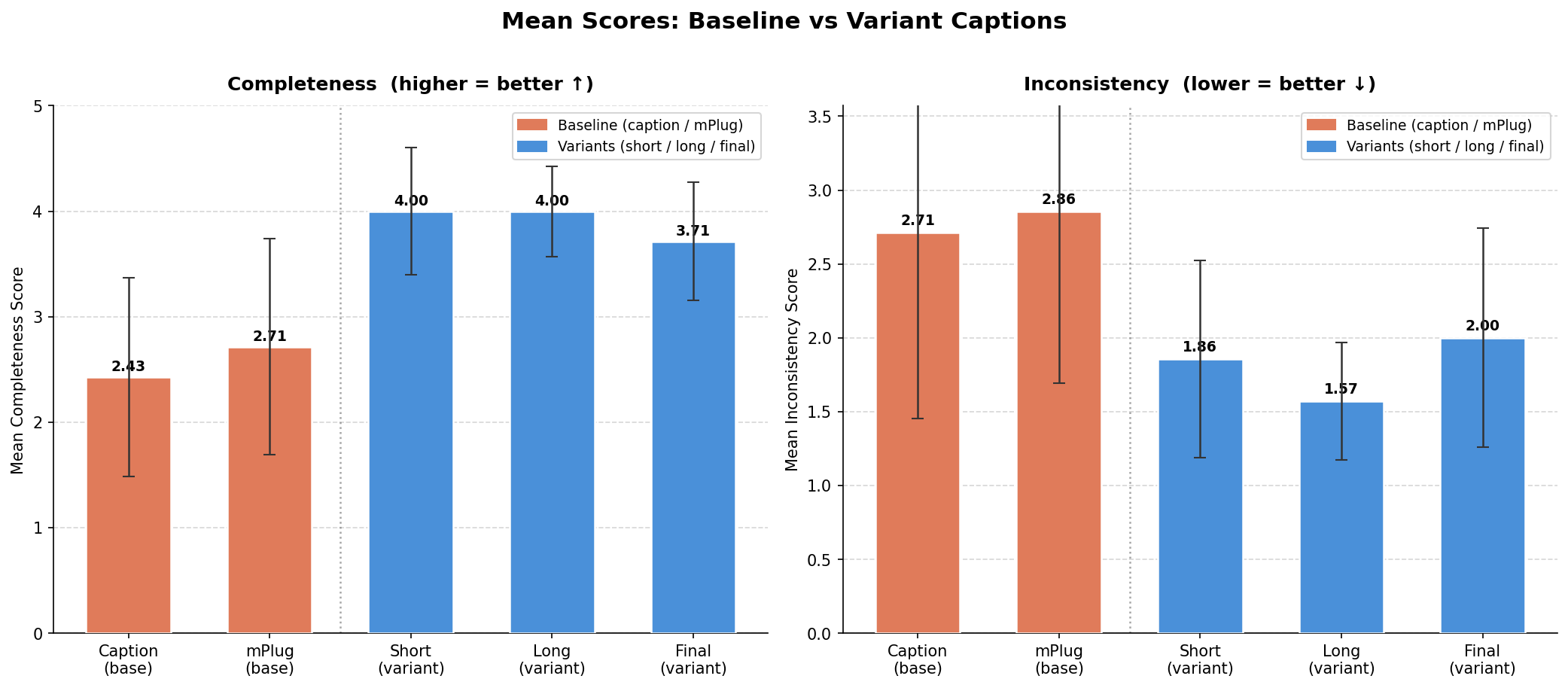}
    \caption{Mean human-evaluated Completeness (left, higher is better) and Inconsistency (right, lower is better) for the two baselines (base caption, mPLUG-2) and three prompt variants (short, long, final), on a 1--5 Likert scale. Error bars denote one standard deviation across annotators and videos.}
    \label{result_mean_bars}
\end{figure*}

To validate whether these improvements translate to perceptual quality, we conducted the human evaluation study described in Section~\ref{sec:human_eval}. Annotators rated the base caption and mPLUG-2 baselines alongside the short, long, and final (iteratively refined) variants on completeness and inconsistency; Fig.~\ref{result_mean_bars} reports the resulting means, and Table~\ref{tab:human_delta} the corresponding relative improvements.

All three variants improve substantially over both baselines on both dimensions. The \textbf{long} variant attains the strongest results overall, improving completeness by \textbf{64.61\%} over the base caption and \textbf{47.60\%} over mPLUG-2, while reducing inconsistency by \textbf{42.07\%} and \textbf{45.10\%} against the same two baselines -- the largest hallucination reduction of any variant. The \textbf{short} variant matches long's completeness gain exactly (both scored 4.00/5) but achieves a more modest \textbf{31.37\%}/\textbf{34.97\%} inconsistency reduction, placing it between long and final on that dimension. The \textbf{final} variant, while not matching the short/long variants' peak completeness, still improves completeness by \textbf{52.67\%} over the base caption and \textbf{36.90\%} over mPLUG-2, and reduces inconsistency by \textbf{26.20\%} and \textbf{30.07\%} against the same baselines. This suggests the iterative refinement stage trades a modest amount of raw object coverage, relative to the single-pass long variant, for a more conservative and concise output, without sacrificing the large margin over both baselines. Notably, despite requiring no training on ground-truth captions, all three variants achieve human preference scores competitive with or superior to mPLUG-2, a fully supervised state-of-the-art captioner underscoring the practical utility of post-hoc, training-free refinement.

\begin{table}[t]
\centering
\caption{Relative improvement ($\Delta$, via Eq.~(3)) in human-evaluated Completeness ($\Delta C$) and Inconsistency ($\Delta I$) for all three variants over each baseline, computed from the printed means in Fig.~\ref{result_mean_bars}.}
\label{tab:human_delta}
\begin{tabular}{lcccc}
\hline
\textbf{Variant} & \multicolumn{2}{c}{\textbf{vs.\ Base Caption}} & \multicolumn{2}{c}{\textbf{vs.\ mPLUG-2}} \\
 & $\Delta C$ & $\Delta I$ & $\Delta C$ & $\Delta I$ \\
\hline
Short & +64.61\% & -31.37\% & +47.60\% & -34.97\% \\
Long & +64.61\% & -42.07\% & +47.60\% & -45.10\% \\
Final & +52.67\% & -26.20\% & +36.90\% & -30.07\% \\
\hline
\end{tabular}
\vspace{2pt}
\end{table}

\subsection{Qualitative Comparison with Large Language Models}
The qualitative analysis presented in Table~\ref{LLM_compare} highlights a key strength of the proposed framework: the generated captions exhibit substantially lower reliance on hallucinated semantics and external prior knowledge while remaining closely grounded in the visual content of the video (Appendix~B presents the complete 10-video comparison) highlights a key strength of the proposed framework: the generated captions exhibit substantially lower reliance on hallucinated semantics and external prior knowledge while remaining more closely grounded in the visual content of the video.. In contrast, Large Language Models such as ChatGPT and Gemini often generate fluent and descriptive captions by leveraging learned world knowledge, which often introduce semantically plausible but visually unsupported information. \inconsistent{Red} spans denote hallucinated or externally inferred content, \consistent{green} spans denote content corroborated across models, and \completed{blue} spans in the proposed model's predictions denote grounded details that both baselines omit. For example, in \textit{video1004}, Gemini identifies the scene as a documentary about Boko Haram and references Mohammed Yusuf and Nigeria, despite these details not being directly observable. Similarly, in \textit{video254}, it attributes the action to a specific football player, whereas the video only depicts a generic football match. ChatGPT exhibits a similar tendency toward abstraction; for instance, in \textit{video106}, it describes performers, spectators, and traditional attire, while the visual content primarily shows children playing outdoors. These examples demonstrate that general-purpose LLMs can be influenced by strong prior associations, whereas the proposed prominence-guided refinement framework remains visually grounded and semantically faithful, and thus more reliable for applications requiring high factual accuracy.

\begin{table*}[htbp]
\centering
\scriptsize
\setlength{\tabcolsep}{3pt}
\renewcommand{\arraystretch}{0.9}
\caption{Comparison of generated captions across different models. \inconsistent{Red} = hallucinated/unsupported; \consistent{Green} = corroborated across captions; \completed{Blue} = grounded detail captured by the proposed model but omitted by both baselines.}
\label{LLM_compare}
\begin{tabularx}{\textwidth}{
l
>{\raggedright\arraybackslash}X
>{\raggedright\arraybackslash}X
>{\raggedright\arraybackslash}X
}
\textbf{Video\_id} & \textbf{ChatGPT} & \textbf{Gemini} & \textbf{Proposed Model Prediction} \\

video106 &
\inconsistent{A performer enthusiastically dances in traditional} \inconsistent{attire on a street while spectators surround him,} \inconsistent{cheering and recording the lively moment.} &
A group of \consistent{children} stand around a patio table outside, singing and using \inconsistent{American Sign Language} to perform. &
The kids are enjoying the summer on a deck, with a person and her \completed{brother} having fun on a \completed{trampoline} at a \completed{birthday party}. \\

video1004 &
The video appears to be a news-style clip showing a shirtless man standing outdoors while \consistent{soldiers in camouflage uniforms} stand behind him. &
This video from \inconsistent{Vice News} details the origins of the militant group \inconsistent{Boko Haram}. It highlights its founder, Muslim cleric \inconsistent{Mohammed Yusuf}. &
A \consistent{soldier} walks past his comrades, then stands guard outside a building, watching as \completed{children walk down the street}. \\

video254 &
A \consistent{football player celebrates passionately after scoring a} dramatic goal during an intense professional match. &
\inconsistent{Cristiano Ronaldo} scores a powerful long-range free kick into the top corner during a professional soccer match. &
The video shows a man's face, he \consistent{celebrates scoring} the \completed{opening} goal with his \completed{team mates}. \\

\end{tabularx}
\end{table*}

\section{Conclusion and Future Work}\label{sec:conclusion}
\label{sec:conclusion}
This work presented ProCap, a post-hoc, model-agnostic framework that combines prominence-guided object ranking with iterative prompt-based refinement to address object omission, hallucination, and semantic incompleteness in video captioning, without retraining the underlying captioning model. Experimental results on the MSR-VTT and MSVD benchmarks validated through object-grounded automatic metrics and human evaluation -- demonstrate consistent improvements in semantic completeness, reduced inconsistency, and stronger alignment with human perception over baseline captioning approaches.

Future work will incorporate complementary modalities such as audio. Further future work plans to employ more robust rectification modules (Appendix~E) to further improve grounding accuracy and reduce repetitive frame context in the caption. We also plan to extend the prominence-guided framework to other multimodal reasoning tasks such as Visual Question Answering, where grounded object representations can enhance both factual correctness and interpretability.


\end{document}